\documentclass[11pt]{article}
\usepackage{acl2016}
\usepackage{times}
\usepackage{latexsym}
\usepackage{url}
\usepackage{verbatim}
\usepackage{multirow}
\usepackage{array}
\usepackage[T1]{fontenc}
\usepackage{textcomp}
\usepackage[utf8]{inputenc}
\usepackage{amssymb}
\usepackage{amsmath}
\usepackage{graphicx}
\usepackage{float}
\usepackage{caption}
\usepackage{subcaption}

\newcolumntype{C}[1]{>{\centering\let\newline\\\arraybackslash\hspace{0pt}}m{#1}}

\aclfinalcopy % Uncomment this line for the final submission
\title{Tweet2Vec: Character-Based Distributed \\
Representations for Social Media}

\author{Bhuwan Dhingra\textsuperscript{1}, Zhong Zhou\textsuperscript{2}, Dylan Fitzpatrick\textsuperscript{1,2}\\{\bf Michael Muehl\textsuperscript{1}} \and {\bf William W. Cohen\textsuperscript{1}}\\
        \textsuperscript{1}School of Computer Science, Carnegie Mellon University, Pittsburgh, PA, USA\\
        \textsuperscript{2}Heinz College, Carnegie Mellon University, Pittsburgh, PA, USA\\
        \texttt{\{bdhingra,djfitzpa,mmuehl\}@andrew.cmu.edu}\\
    \texttt{zhongzhou@cmu.edu} \quad \texttt{wcohen@cs.cmu.edu}}

\date{}

\begin{document}

\maketitle

\begin{abstract}
    Text from social media provides a set of challenges that can cause traditional NLP approaches to fail. Informal language, spelling errors, abbreviations, and special characters are all commonplace in these posts, leading to a prohibitively large vocabulary size for word-level approaches. We propose a character composition model, \texttt{tweet2vec}, which finds vector-space representations of whole tweets by learning complex, non-local dependencies in character sequences. The proposed model outperforms a word-level baseline at predicting user-annotated \textit{hashtags} associated with the posts, doing significantly better when the input contains many out-of-vocabulary words or unusual character sequences. Our \textit{tweet2vec} encoder is publicly available\footnote{https://github.com/bdhingra/tweet2vec}.
\end{abstract}

\section{Introduction}
We understand from Zipf's Law that in any natural language corpus a majority of the vocabulary word types will either be absent or occur in low frequency. Estimating the statistical properties of these rare word types is naturally a difficult task. This is analogous to the curse of dimensionality when we deal with sequences of tokens - most sequences will occur only once in the training data. Neural network architectures overcome this problem by defining non-linear compositional models over vector space representations of tokens and hence assign non-zero probability even to sequences not seen during training \cite{bengio2003neural,kiros2015skip}. In this work, we explore a similar approach to learning distributed representations of social media posts by composing them from their constituent \textit{characters}, with the goal of generalizing to out-of-vocabulary words as well as sequences at test time.

Traditional Neural Network Language Models (NNLMs) treat words as the basic units of language and assign independent vectors to each word type. To constrain memory requirements, the vocabulary size is fixed before-hand; therefore, rare and out-of-vocabulary words are all grouped together under a common type `UNKNOWN'. This choice is motivated by the assumption of arbitrariness in language, which means that surface forms of words have little to do with their semantic roles. Recently, \cite{ling2015finding} challenge this assumption and present a bidirectional Long Short Term Memory (LSTM) \cite{hochreiter1997long} for composing word vectors from their constituent characters which can memorize the arbitrary aspects of word orthography as well as generalize to rare and out-of-vocabulary words.

Encouraged by their findings, we extend their approach to a much larger unicode character set, and model long sequences of text as functions of their constituent characters (including white-space). We focus on social media posts from the website Twitter, which are an excellent testing ground for character based models due to the noisy nature of text. Heavy use of slang and abundant misspellings means that there are many orthographically and semantically similar tokens, and special characters such as emojis are also immensely popular and carry useful semantic information. In our moderately sized training dataset of 2 million tweets, there were about 0.92 million unique word types. It would be expensive to capture all these phenomena in a word based model in terms of both the memory requirement (for the increased vocabulary) and the amount of training data required for effective learning. Additional benefits of the character based approach include language independence of the methods, and no requirement of NLP preprocessing such as word-segmentation.

A crucial step in learning good text representations is to choose an appropriate objective function to optimize. Unsupervised approaches attempt to reconstruct the original text from its latent representation \cite{mikolov2013efficient,bengio2003neural}. Social media posts however, come with their own form of supervision annotated by millions of users, in the form of \textit{hashtags} which link posts about the same topic together. A natural assumption is that the posts with the same hashtags should have embeddings which are close to each other. Hence, we formulate our training objective to maximize cross-entropy loss at the task of predicting hashtags for a post from its latent representation. 

We propose a Bi-directional Gated Recurrent Unit (Bi-GRU) \cite{chung2014empirical} neural network for learning tweet representations. Treating white-space as a special character itself, the model does a forward and backward pass over the entire sequence, and the final GRU states are linearly combined to get the tweet embedding. Posterior probabilities over hashtags are computed by projecting this embedding to a softmax output layer. Compared to a word-level baseline this model shows improved performance at predicting hashtags for a held-out set of posts. Inspired by recent work in learning vector space text representations, we name our model \textit{tweet2vec}. 

\section{Related Work}
Using neural networks to learn distributed representations of words dates back to \cite{bengio2003neural}. More recently, \cite{mikolov2013efficient} released \texttt{word2vec} - a collection of word vectors trained using a recurrent neural network. These word vectors are in widespread use in the NLP community, and the original work has since been extended to sentences \cite{kiros2015skip}, documents and paragraphs \cite{le2014distributed}, topics \cite{niu2015topic2vec} and queries \cite{grbovic2015context}. All these methods require storing an extremely large table of vectors for all word types and cannot be easily generalized to unseen words at test time \cite{ling2015finding}. They also require preprocessing to find word boundaries which is non-trivial for a social network domain like Twitter. 

In \cite{ling2015finding}, the authors present a compositional character model based on bidirectional LSTMs as a potential solution to these problems. A major benefit of this approach is that large word lookup tables can be compacted into character lookup tables and the compositional model scales to large data sets better than other state-of-the-art approaches. While \cite{ling2015finding} generate word embeddings from character representations, we propose to generate vector representations of entire tweets from characters in our \texttt{tweet2vec} model.

Our work adds to the growing body of work showing the applicability of character models for a variety of NLP tasks such as Named Entity Recognition \cite{santos2015boosting}, POS tagging \cite{santos2014learning}, text classification \cite{zhang2015character} and language modeling \cite{karpathy2015visualizing,kim2015character}.

Previously, \cite{luong2013better} dealt with the problem of estimating rare word representations by building them from their constituent morphemes. While they show improved performance over word-based models, their approach requires a morpheme parser for preprocessing which may not perform well on noisy text like Twitter. Also the space of all morphemes, though smaller than the space of all words, is still large enough that modelling all morphemes is impractical.

Hashtag prediction for social media has been addressed earlier, for example in \cite{weston2014tagspace,godin2013using}. \cite{weston2014tagspace} also use a neural architecture, but compose text embeddings from a lookup table of words. They also show that the learned embeddings can generalize to an unrelated task of document recommendation, justifying the use of hashtags as supervision for learning text representations.

\section{Tweet2Vec}
\begin{figure}[h]
\begin{center}
\fbox{\includegraphics[width=0.8\columnwidth]{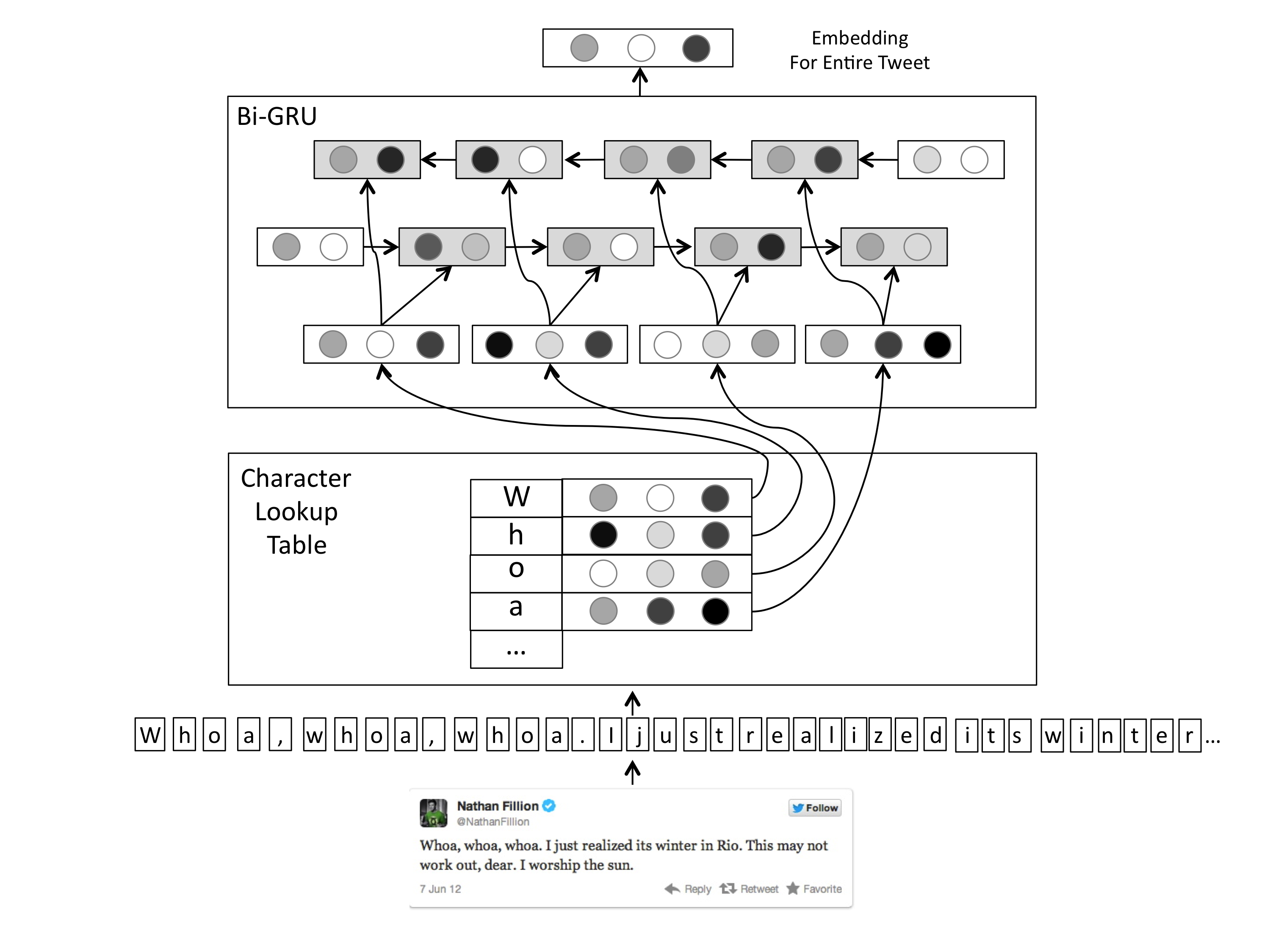}}
\end{center}
\caption{Tweet2Vec encoder for social media text}
\label{fig:1layer}
\end{figure}

\textbf{Bi-GRU Encoder:} \quad
Figure \ref{fig:1layer} shows our model for encoding tweets. It uses a similar structure to the C2W model in \cite{ling2015finding}, with LSTM units replaced with GRU units. 

The input to the network is defined by an alphabet of characters $C$ (this may include the entire unicode character set). The input tweet is broken into a stream of characters $c_1, c_2, ... c_m$ each of which is represented by a $1$-by-$|C|$ encoding. These one-hot vectors are then projected to a character space by multiplying with the matrix $P_C \in \mathbb{R}^{|C| \times d_c}$, where $d_c$ is the dimension of the character vector space. Let $x_1, x_2, ... x_m$ be the sequence of character vectors for the input tweet after the lookup. The encoder consists of a forward-GRU and a backward-GRU. Both have the same architecture, except the backward-GRU processes the sequence in reverse order. Each of the GRU units process these vectors sequentially, and starting with the initial state $h_0$ compute the sequence $h_1, h_2, ... h_m$ as follows: 
\begin{align*}
r_t &= \sigma (W_r x_t + U_r h_{t-1} + b_r), \\
z_t &= \sigma (W_z x_t + U_z h_{t-1} + b_z), \\
\tilde{h}_t &= tanh(W_h x_t + U_h (r_t \odot h_{t-1}) + b_h), \\
h_t &= (1-z_t) \odot h_{t-1} + z_t \odot \tilde{h}_t.
\end{align*}

Here $r_t$, $z_t$ are called the \textit{reset} and \textit{update} gates respectively, and $\tilde{h}_t$ is the \textit{candidate} output state which is converted to the actual output state $h_t$. $W_r, W_z, W_h$ are $d_h \times d_c$ matrices and $U_r, U_z, U_h$ are $d_h \times d_h$ matrices, where $d_h$ is the hidden state dimension of the GRU. The final states $h_m^f$ from the forward-GRU, and $h_0^b$ from the backward GRU are combined using a fully-connected layer to the give the final tweet embedding $e_t$:
\begin{equation}\label{eq:1}
e_t = W^f h_m^f + W^b h_0^b
\end{equation}
Here $W^f, W^b$ are $d_t \times d_h$ and $b$ is $d_t \times 1$ bias term, where $d_t$ is the dimension of the final tweet embedding. In our experiments we set $d_t=d_h$. All parameters are learned using gradient descent.

\textbf{Softmax:} \quad
Finally, the tweet embedding is passed through a linear layer whose output is the same size as the number of hashtags $L$ in the data set. We use a softmax layer to compute the posterior hashtag probabilities:
\begin{equation}\label{eq:2}
P(y=j |e) = \frac{exp(w_j^Te + b_j)}{\sum_{i=1}^L exp(w_i^Te + b_j)}. 
\end{equation}

%\subsection{2-Layer (\textit{char2word2vec})}
%While GRU units can theoretically possess arbitrarily long memories, learning short term dependencies is usually easier \cite{sutskever2014sequence}. Intuitively, it also seems that while words may depend on each other, the characters within a word should not depend on characters from another word. With this motivation, we have implemented a hierarchical version of the model described above, which composes word representations from characters in the first layer and tweet representations from the words in the second layer. The input text is split into words by tokenizing along white-spaces. Weights are shared between the word encoders in the first layer for different words. Figure \ref{fig:2layer} illustrates this model, which is essentially a stacked version of the model described in the previous version. The sequence lengths are typically much shorter than in the single-layer model, as sequences now correspond to characters in a word or words in a tweet instead of characters in a tweet, which has the added benefit of faster training. The final tweet embedding is passed through the softmax layer described in equation \ref{eq:2}, and a prediction is obtained by taking the max over the computed probabilities.
%
%\begin{figure}[h]
%\begin{center}
%\fbox{\includegraphics[width=\columnwidth]{2layer_fig}}
%\end{center}
%\caption{Char2Word2Vec: two-layer tweet encoder. Square boxes represent vectors of neuron activations. Shaded boxes indicate non-linearities.}
%\label{fig:2layer}
%\end{figure}

\textbf{Objective Function:} \quad
We optimize the categorical cross-entropy loss between predicted and true hashtags: 
\begin{equation}
    J = \frac{1}{B} \sum_{i=1}^{B} \sum_{j=1}^{L} -t_{i,j}log(p_{i,j}) + \lambda \|\Theta\|^2.
\end{equation}
Here $B$ is the batch size, $L$ is the number of classes, $p_{i,j}$ is the predicted probability that the $i$-th tweet has hashtag $j$, and $t_{i,j} \in \{0,1\}$ denotes the ground truth of whether the $j$-th hashtag is in the $i$-th tweet. We use L2-regularization weighted by $\lambda$.

\section{Experiments and Results}
\subsection{Word Level Baseline}
Since our objective is to compare character-based and word-based approaches, we have also implemented a simple word-level encoder for tweets. The input tweet is first split into tokens along white-spaces. A more sophisticated tokenizer may be used, but for a fair comparison we wanted to keep language specific preprocessing to a minimum. The encoder is essentially the same as \textit{tweet2vec}, with the input as words instead of characters. A lookup table stores word vectors for the $V$ (20K here) most common words, and the rest are grouped together under the `UNK' token. 

\subsection{Data}
Our dataset consists of a large collection of global posts from Twitter\footnote{\url{https://twitter.com/}} between the dates of June 1, 2013 to June 5, 2013. Only English language posts (as detected by the \textit{lang} field in Twitter API) and posts with at least one hashtag are retained. We removed infrequent hashtags ($<500$ posts) since they do not have enough data for good generalization. We also removed very frequent tags ($>19K$ posts) which were almost always from automatically generated posts (ex: \texttt{\#androidgame}) which are trivial to predict. The final dataset contains 2 million tweets for training, 10K for validation and 50K for testing, with a total of 2039 distinct hashtags. We use simple regex to preprocess the post text and remove hashtags (since these are to be predicted) and HTML tags, and replace user-names and URLs with special tokens. We also removed \textit{retweets} and convert the text to lower-case.

%\begin{table}
%\centering
%\begin{tabular}{|c|c|}
%\hline
%\# Hashtags & 2039 \\
%Training & 2M \\
%Validation & 250K \\
%Test & 50K\\
%\hline
%\end{tabular}
%\caption{Number of hashtags and data sizes}
%\label{tab:data}
%\end{table}

\begin{table*}[!t]
\small
\centering
\begin{tabular}{|m{8cm}|m{3.6cm}|m{3.1cm}|}
\hline
\textbf{Tweets} & \textbf{Word model baseline} & \textbf{\textit{tweet2vec}}\\
\hline
\texttt{ninety-one degrees.\includegraphics[scale=0.4]{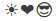}} & \texttt{\#initialsofsomeone.. \#nw \#gameofthrones} & \texttt{\#summer \textbf{\#loveit} \#sun}\\
\hline
\texttt{self-cooked scramble egg. yum!!     !url} & \texttt{\#music \#cheap \#cute} & \texttt{\textbf{\#yummy} \#food \#foodporn}\\
\hline
\texttt{can't sleeeeeeep} & \texttt{\#gameofthrones \#heartbreaker} & \texttt{\#tired \textbf{\#insomnia}}\\
\hline
\texttt{oklahoma!!!!!!!!!!! champions!!!!!} & \texttt{\#initialsofsomeone.. \#nw \#lrt} & \texttt{\textbf{\#wcws} \#sooners \#ou}\\
\hline
\texttt{7 \% of battery . iphones die too quick .} & \texttt{\#help \#power \#money \#s} & \texttt{\#fml \#apple \#bbl \textbf{\#thestruggle}}\\
\hline
\texttt{i have the cutest nephew in the world   !url} & \texttt{\#nephew \textbf{\#cute} \#family} & \texttt{\#socute \textbf{\#cute} \#puppy}\\
\hline
%\texttt{100 pushups, 100 tricep dips, 200 v ups and 100 calf raises.  nice late night workout} & \texttt{\#music \#golf \#amazon} & \texttt{\#fitness \#fb \#workout}\\
%\hline
\end{tabular}
\caption{Examples of top predictions from the models. The correct hashtag(s) if detected are in bold.}
\label{tab:ex}
\end{table*}

\subsection{Implementation Details}
Word vectors and character vectors are both set to size $d_L=150$ for their respective models. There were 2829 unique characters in the training set and we model each of these independently in a character look-up table. Embedding sizes were chosen such that each model had roughly the same number of parameters (Table \ref{tab:size}). Training is performed using mini-batch gradient descent with Nesterov's momentum. We use a batch size $B=64$, initial learning rate $\eta_0=0.01$ and momentum parameter $\mu_0=0.9$. L2-regularization with $\lambda=0.001$ was applied to all models. Initial weights were drawn from 0-mean gaussians with $\sigma=0.1$ and initial biases were set to 0. The hyperparameters were tuned one at a time keeping others fixed, and values with the lowest validation cost were chosen. The resultant combination was used to train the models until performance on validation set stopped increasing. During training, the learning rate is halved everytime the validation set precision increases by less than 0.01 \% from one epoch to the next. The models converge in about 20 epochs. Code for training both the models is publicly available on github.

\begin{table}
    \centering
    \begin{tabular}{m{3.7cm}|m{1.2cm}|m{1.2cm}}
        & word & \textit{tweet2vec} \\
        \hline
        $d_t$, $d_h$ & 200 & 500 \\
        Total Parameters & 3.91M & 3.90M \\
        Training Time / Epoch & \textbf{1528s} & 9649s \\
    \end{tabular}
    \caption{Model sizes and training time/epoch}
    \label{tab:size}
\end{table}

%\begin{table*}
%\centering
%\begin{tabular}{|c|c|c|c|}
%\hline
%Model & Number of hidden units & Total Parameters & Training Time for 1 Epoch \\
%\hline
%word & 200 & 3.91M & \textbf{1528s} \\
%\textit{\textit{char2vec}} & 500 & 3.90M & 9649s \\
%\textit{char2word2vec} & 320,350 & 3.90M & 5838s \\
%\hline
%\end{tabular}
%\caption{Model sizes and training time/epoch}
%\label{tab:size}
%\end{table*}

\subsection{Results}
We test the character and word-level variants by predicting hashtags for a held-out test set of posts. Since there may be more than one correct hashtag per post, we generate a ranked list of tags for each post from the output posteriors, and report average precision@1, recall@10 and mean rank of the correct hashtags. These are listed in Table \ref{tab:res}.

\begin{table}
\centering
\begin{tabular}{|C{1.6cm}|C{1.5cm}|C{1.5cm}|C{1.4cm}|}
\hline
Model & Precision @1 & Recall @10 & Mean Rank \\
\hline
\hline
\multicolumn{4}{|c|}{Full test set (50K)}\\
\hline
word& 24.1\% & 42.8\% & 133\\
\textit{tweet2vec}& \textbf{28.4\%} & \textbf{48.5\%} & \textbf{104}\\
\hline
\multicolumn{4}{|c|}{Rare words test set (2K)}\\
\hline
word& 20.4\% & 37.2\% & 167\\
\textit{tweet2vec}& \textbf{32.9\%} & \textbf{51.3\%} & \textbf{104}\\
\hline
\multicolumn{4}{|c|}{Frequent words test set (2K)}\\
\hline
word& 20.9\% & 41.3\% & 133\\
\textit{tweet2vec}& \textbf{23.9\%} & \textbf{44.2\%} & \textbf{112}\\
%\textit{char2word2vec} & 12.8\% & 26.5\% & 288\\
\hline
\end{tabular}
\caption{Hashtag prediction results. Best numbers for each test set are in bold.}
\label{tab:res}
\end{table}

\begin{figure*}
    \centering
    \begin{subfigure}[b]{0.3\textwidth}
        \includegraphics[width=\textwidth]{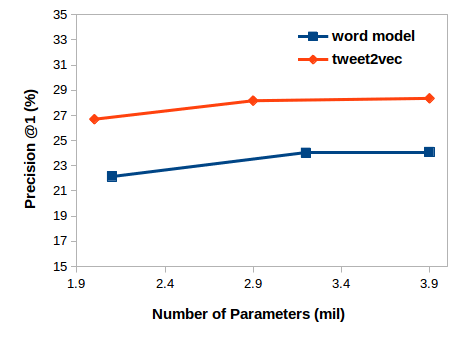}
        \caption{Full Test Set}
        \label{fig:full}
    \end{subfigure}
    ~ %add desired spacing between images, e. g. ~, \quad, \qquad, \hfill etc. 
    \begin{subfigure}[b]{0.3\textwidth}
        \includegraphics[width=\textwidth]{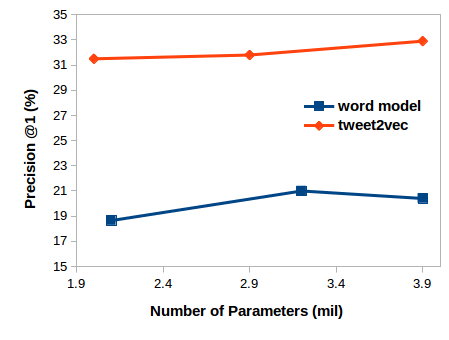}
        \caption{Rare Words Test Set}
        \label{fig:rare}
    \end{subfigure}
    ~ %add desired spacing between images, e. g. ~, \quad, \qquad, \hfill etc. 
    \begin{subfigure}[b]{0.3\textwidth}
        \includegraphics[width=\textwidth]{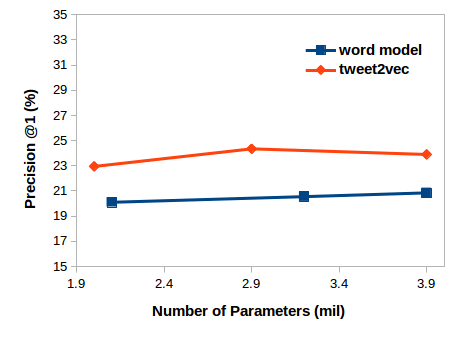}
        \caption{Frequent Words Test Set}
        \label{fig:freq}
    \end{subfigure}
    \caption{Precision @1 v Number of model parameters for word model and \textit{tweet2vec}.}\label{fig:p}
\end{figure*}

To see the performance of each model on posts containing rare words (RW) and frequent words (FW) we selected two test sets each containing 2,000 posts. We populated these sets with posts which had the maximum and minimum number of out-of-vocabulary words respectively, where vocabulary is defined by the 20K most frequent words. Overall, \textit{tweet2vec} outperforms the word model, doing significantly better on RW test set and comparably on FW set. This improved performance comes at the cost of increased training time (see Table \ref{tab:size}), since moving from words to characters results in longer input sequences to the GRU. 

We also study the effect of model size on the performance of these models. For the word model we set vocabulary size $V$ to 8K, 15K and 20K respectively. For \textit{tweet2vec} we set the GRU hidden state size to 300, 400 and 500 respectively. Figure \ref{fig:p} shows precision \@1 of the two models as the number of parameters is increased, for each test set described above. There is not much variation in the performance, and moreover \textit{tweet2vec} always outperforms the word based model for the same number of parameters.

\begin{table}
\centering
\begin{tabular}{|c|c|c|c|}
\hline
Dataset & \# Hashtags & word & \textit{tweet2vec}\\
\hline
\hline
small & 933 & 28.0\% & \textbf{33.1}\%\\
\hline
medium & 2039 & 24.1\% & \textbf{28.4}\%\\
\hline
large & 5114 & 20.1\% & \textbf{24.6}\%\\
\hline
\end{tabular}
\caption{Precision @1 as training data size and number of output labels is increased. Note that the test set is different for each setting.}
\label{tab:numtags}
\end{table}

Table \ref{tab:numtags} compares the models as complexity of the task is increased. We created 3 datasets (small, medium and large) with an increasing number of hashtags to be predicted. This was done by varying the lower threshold of the minimum number of tags per post for it to be included in the dataset. Once again we observe that \textit{tweet2vec} outperforms its word-based counterpart for each of the three settings. 

%The single layer \textit{char2vec} outperformed its two-layer counterpart on all three test sets. In general, deeper networks are more difficult to train than shallow ones and require a much more careful tuning of hyperparameters \cite{sutskever2013importance}. We note that it may be possible to improve the performance of the \textit{char2word2vec} model with better initialization and training. Also, faster training (see table \ref{tab:size}) of the two-layer model compared to 1-layer makes it suitable when speed is important. 

Finally, table \ref{tab:ex} shows some predictions from the word level model and \textit{tweet2vec}. We selected these to highlight some strengths of the character based approach - it is robust to word segmentation errors and spelling mistakes, effectively interprets emojis and other special characters to make predictions, and also performs comparably to the word-based approach for in-vocabulary tokens.

\section{Conclusion}
We have presented \textit{tweet2vec} - a character level encoder for social media posts trained using supervision from associated \textit{hashtags}. Our result shows that \textit{tweet2vec} outperforms the word based approach, doing significantly better when the input post contains many rare words. We have focused only on English language posts, but the character model requires no language specific preprocessing and can be extended to other languages. For future work, one natural extension would be to use a character-level decoder for predicting the hashtags. This will allow generation of hashtags not seen in the training dataset. Also, it will be interesting to see how our tweet2vec embeddings can be used in domains where there is a need for semantic understanding of social media, such as tracking infectious diseases \cite{signorini2011use}. Hence, we provide an off-the-shelf encoder trained on \textit{medium} dataset described above to compute vector-space representations of tweets along with our code on github.

%There are many immediate future research questions to be explored and answered. 
%
%On the theory side, since each tweet is composed of one or more sentences. Possible extension of the two-layer LSTM to multiple layers involving sentences or n-grams would be interesting. However, these extensions may not help performance as tweet is a special kind of communication based on short text. It is also interesting to use theory on provable bounds to help the deep networks to converge faster. 
%
%We are also looking forward to advances on the engineering side, the 1-layer character-level model is extremely hard to train as the number of hyper-parameters grows. Better hyper-parameter tuning with larger and deeper neural networks is really what we will continue to engineer on. It may give more interesting result. 

\section*{Acknowledgments}
We would like to thank Alex Smola, Yun Fu, Hsiao-Yu Fish Tung, Ruslan Salakhutdinov, and Barnabas Poczos for useful discussions. We would also like to thank Juergen Pfeffer for providing access to the Twitter data, and the reviewers for their comments.

\bibliography{biblio}

\begin{thebibliography}{}

\bibitem[\protect\citename{Bengio \bgroup et al.\egroup
  }2003]{bengio2003neural}
Yoshua Bengio, R{\'e}jean Ducharme, Pascal Vincent, and Christian Janvin.
\newblock 2003.
\newblock A neural probabilistic language model.
\newblock {\em The Journal of Machine Learning Research}, 3:1137--1155.

\bibitem[\protect\citename{Chung \bgroup et al.\egroup
  }2014]{chung2014empirical}
Junyoung Chung, Caglar Gulcehre, KyungHyun Cho, and Yoshua Bengio.
\newblock 2014.
\newblock Empirical evaluation of gated recurrent neural networks on sequence
  modeling.
\newblock {\em arXiv preprint arXiv:1412.3555}.

\bibitem[\protect\citename{Godin \bgroup et al.\egroup }2013]{godin2013using}
Fr{\'e}deric Godin, Viktor Slavkovikj, Wesley De~Neve, Benjamin Schrauwen, and
  Rik Van~de Walle.
\newblock 2013.
\newblock Using topic models for twitter hashtag recommendation.
\newblock In {\em Proceedings of the 22nd international conference on World
  Wide Web companion}, pages 593--596. International World Wide Web Conferences
  Steering Committee.

\bibitem[\protect\citename{Grbovic \bgroup et al.\egroup
  }2015]{grbovic2015context}
Mihajlo Grbovic, Nemanja Djuric, Vladan Radosavljevic, Fabrizio Silvestri, and
  Narayan Bhamidipati.
\newblock 2015.
\newblock Context-and content-aware embeddings for query rewriting in sponsored
  search.
\newblock In {\em Proceedings of the 38th International ACM SIGIR Conference on
  Research and Development in Information Retrieval}, pages 383--392. ACM.

\bibitem[\protect\citename{Hochreiter and Schmidhuber}1997]{hochreiter1997long}
Sepp Hochreiter and J{\"u}rgen Schmidhuber.
\newblock 1997.
\newblock Long short-term memory.
\newblock {\em Neural computation}, 9(8):1735--1780.

\bibitem[\protect\citename{Karpathy \bgroup et al.\egroup
  }2015]{karpathy2015visualizing}
Andrej Karpathy, Justin Johnson, and Fei-Fei Li.
\newblock 2015.
\newblock Visualizing and understanding recurrent networks.
\newblock {\em arXiv preprint arXiv:1506.02078}.

\bibitem[\protect\citename{Kim \bgroup et al.\egroup }2015]{kim2015character}
Yoon Kim, Yacine Jernite, David Sontag, and Alexander~M Rush.
\newblock 2015.
\newblock Character-aware neural language models.
\newblock {\em arXiv preprint arXiv:1508.06615}.

\bibitem[\protect\citename{Kiros \bgroup et al.\egroup }2015]{kiros2015skip}
Ryan Kiros, Yukun Zhu, Ruslan Salakhutdinov, Richard~S Zemel, Antonio Torralba,
  Raquel Urtasun, and Sanja Fidler.
\newblock 2015.
\newblock Skip-thought vectors.
\newblock {\em arXiv preprint arXiv:1506.06726}.

\bibitem[\protect\citename{Le and Mikolov}2014]{le2014distributed}
Quoc~V Le and Tomas Mikolov.
\newblock 2014.
\newblock Distributed representations of sentences and documents.
\newblock {\em arXiv preprint arXiv:1405.4053}.

\bibitem[\protect\citename{Ling \bgroup et al.\egroup }2015]{ling2015finding}
Wang Ling, Tiago Lu{\'\i}s, Lu{\'\i}s Marujo, Ram{\'o}n~Fernandez Astudillo,
  Silvio Amir, Chris Dyer, Alan~W Black, and Isabel Trancoso.
\newblock 2015.
\newblock Finding function in form: Compositional character models for open
  vocabulary word representation.
\newblock {\em arXiv preprint arXiv:1508.02096}.

\bibitem[\protect\citename{Luong \bgroup et al.\egroup }2013]{luong2013better}
Thang Luong, Richard Socher, and Christopher~D Manning.
\newblock 2013.
\newblock Better word representations with recursive neural networks for
  morphology.
\newblock In {\em CoNLL}, pages 104--113. Citeseer.

\bibitem[\protect\citename{Mikolov \bgroup et al.\egroup
  }2013]{mikolov2013efficient}
Tomas Mikolov, Kai Chen, Greg Corrado, and Jeffrey Dean.
\newblock 2013.
\newblock Efficient estimation of word representations in vector space.
\newblock {\em arXiv preprint arXiv:1301.3781}.

\bibitem[\protect\citename{Niu and Dai}2015]{niu2015topic2vec}
Li-Qiang Niu and Xin-Yu Dai.
\newblock 2015.
\newblock Topic2vec: Learning distributed representations of topics.
\newblock {\em arXiv preprint arXiv:1506.08422}.

\bibitem[\protect\citename{Santos and Guimar{\~a}es}2015]{santos2015boosting}
Cicero Nogueira~dos Santos and Victor Guimar{\~a}es.
\newblock 2015.
\newblock Boosting named entity recognition with neural character embeddings.
\newblock {\em arXiv preprint arXiv:1505.05008}.

\bibitem[\protect\citename{Santos and Zadrozny}2014]{santos2014learning}
Cicero~D Santos and Bianca Zadrozny.
\newblock 2014.
\newblock Learning character-level representations for part-of-speech tagging.
\newblock In {\em Proceedings of the 31st International Conference on Machine
  Learning (ICML-14)}, pages 1818--1826.

\bibitem[\protect\citename{Signorini \bgroup et al.\egroup
  }2011]{signorini2011use}
Alessio Signorini, Alberto~Maria Segre, and Philip~M Polgreen.
\newblock 2011.
\newblock The use of twitter to track levels of disease activity and public
  concern in the us during the influenza a h1n1 pandemic.
\newblock {\em PloS one}, 6(5):e19467.

\bibitem[\protect\citename{Weston \bgroup et al.\egroup
  }2014]{weston2014tagspace}
Jason Weston, Sumit Chopra, and Keith Adams.
\newblock 2014.
\newblock tagspace: Semantic embeddings from hashtags.
\newblock In {\em Proceedings of the 2014 Conference on Empirical Methods in
  Natural Language Processing (EMNLP)}, pages 1822--1827.

\bibitem[\protect\citename{Zhang \bgroup et al.\egroup
  }2015]{zhang2015character}
Xiang Zhang, Junbo Zhao, and Yann LeCun.
\newblock 2015.
\newblock Character-level convolutional networks for text classification.
\newblock In {\em Advances in Neural Information Processing Systems}, pages
  649--657.

\end{thebibliography}
\bibliographystyle{acl2016}

%\appendix
%\section{More Examples}
%\begin{table*}
%\small
%\centering
%\begin{tabular}{|m{8.3cm}|m{3.3cm}|m{3.3cm}|}
%\hline
%\textbf{Posts from Rare words test set} & \textbf{Word model baseline} & \textbf{\textit{tweet2vec}}\\
%\hline
%\texttt{c'mon rafa!!!!!} & \texttt{\#fb \#music \#follow} & \texttt{\#rg13 \#fb \textbf{\#frenchopen}}\\
%\hline
%\texttt{yuchun, yuchunnie, chunnie, chun, chunah....ahhhh to many pyc cute names...} & \texttt{\#win \#fb \#asco13} & \texttt{\#28thyuchunday \textbf{\#happy28thpark- yuchun} \#happyhyunjoongday}\\
%\hline
%\texttt{:- psychologysays, the desire toosleep a lot mayreveal feelings ofloneliness,anxiousness and ordepression.} & \texttt{\#asco13 \#foursquare \#cancer} & \texttt{\textbf{\#fact} \#english \#fail}\\
%\hline
%\texttt{@user yaaaaaaaaaaaaaaaaaaaaaaaaaaaaaaaaaaa were readyyyy} & \texttt{\#fb \#lol \#justsaying} & \texttt{\#excited \#lol \#cantwait}\\
%\hline
%\texttt{j.b n.h z.m h.s l.p l.t n.c} & \texttt{\#fb \#c \#follow} & \texttt{\textbf{\#initialsof- someonespecial} \#r \#middleschool- memories}\\
%\hline
%\end{tabular}
%\caption{Examples of top hashtags predicted by the two models. The correct hashtag(s), if detected, are in bold.}
%\label{tab:app}
%\end{table*}

\end{document}